# Post-Earthquake Assessment of Buildings Using Deep Learning

Dhananjay Nahata[1*], Harish Kumar Mulchandani[1], Suraj Bansal[1], and G. Muthukumar[2]

[1]Birla Institute of Technology & Science, Pilani, India
f2015812@pilani.bits-pilani.ac.in
f2011451@pilani.bits-pilani.ac.in

**Abstract.**

Classification of the extent of damage suffered by a building in a seismic event is crucial from the safety perspective and repairing work. In this study, authors have proposed a CNN based autonomous damage detection model. Over 1200 images of different types of buildings-1000 for training and 200 for testing classified into 4 categories according to the extent of damage suffered. Categories are namely, no damage, minor damage, major damage, and collapse. Trained network tested by the application of various algorithms with different learning rates. The most optimum results were obtained on the application of VGG16 transfer learning model with a learning rate of 1e-5 as it gave a training accuracy of 97.85% and validation accuracy of up to 89.38%. The model developed has real-time application in the event of an earthquake.

## 1    Introduction

In recent years, the aging of large-scale structures has become a hot burning area of research in the field of Civil and Structural Engineering especially for low rise buildings, thus performing an irreplaceable function in present-day society. From this, we can ensure structural stability and detecting structural damage which has gained attention by researchers in this field. Traditionally Structural Health Monitoring (SHM) approaches require dense data to predict the structural damage of any structure and to measure those damages many inspection instruments and various other sensors and machines are used to detect the defects and damages through manual inspection. Online monitoring of structures is required for estimation of the remaining useful life of the structure. This process requires updating of structural parameters using new information obtained through response quantities. For SHM, computer-vision based techniques have been developed, and recent techniques involve usage of Deep Learning techniques.

At various present authors deep learning, technologies have been applied in various civil engineering application such as building assessment [1], Crack damage detection and damage[2-6], Real-time traffic management [7],  Seismic reliability [8], Pixel level detection for structures and pavements[9-10]. State of the art in CNN's Krizhevsky et al. created AlexNet[11], GoogLeNet by Szegedy et al. [12] and VGG16 by Simonyan & Zisserman [13]. Such CNN Net could be applied using transfer learning for the new database, Gao and Mosalam [5] applied the transfer learning using VGG16 on the image based structural recognition. Various authors have used segmentation techniques for damage detection and other purposes [14-15].

## 2    Methodology

### 2.1    Convolutional Neural Network Design

These are the core layers of the CNN architecture which we used on our datasets. It performs three operations on any input array which we have given for identification and classification. Firstly, it performs multiplications (i.e., dot product) element – by - element between a sub-array of an input array and their corresponding receptive field. Sometimes Receptive fields are also called as filters or kernels. The initial weight values of these filters or kernels are randomly generated. The values of the biases can be set accordingly in many different ways in correspondence with the network configurations. One of the most well-known configurations for initializations of the biases is found from Krizhevsky (2012). Both values are tuned in training using Stochastic Gradient Descent algorithms. The size of subarray is always equal to the receptive field, but the receptive field is always smaller than the shape of the input array. Secondly, we sum up the multiplied values and add biases to those summed values.  The most significant advantage we have by applying a convolutional layer is that it reduces the size of the input which indirectly reduces our computational cost, thus making the classification process, time as well as cost economical. The additional hyper-parameter of this layer is stride.  The stride defines sliding of the receptive field's columns and rows (pixels) during a particular instant across the input array's width and height. The number of filters to be used is also a parameter defined. The number of filters used represents the depth of the output layers, i.e. the third dimension of the output layer. A larger stride size leads to very less receptive field applications and smaller output size, which might lead to a loss in data. However, the advantage of using larger strides is that computational cost reduces drastically, making the process more cost economic. The output size of the convolutional layer is calculated through the following equation:

$$O_n = \frac{(I_n - P_n)}{S_n} + 1 \qquad (1)$$

Where $O_n$ = Output size of the nth layer
$I_n$ = Input size of nth layer,  $P_n$ = Pooling size of nth layer,  $S_n$ = Stride size of the nth layer

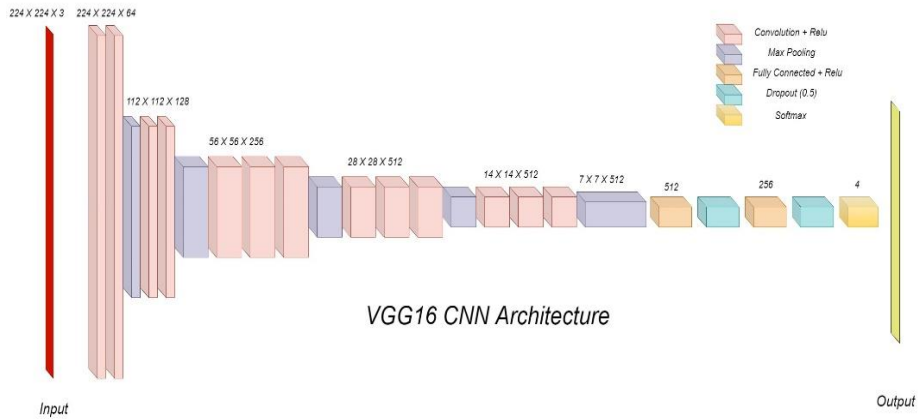

Figure 1: VGG16 CNN Architecture

## 2.2 Pooling Layers

Another key aspect of CNN is a pooling layer, which reduces the spatial size of the input array, thus reducing the computational cost and making computations faster [2]. This process is also named as downsampling. Two different pooling operations are defined. Max pooling takes maximum value from the subarrays of the corresponding input arrays whereas mean pooling takes the mean of the subarrays of the input arrays. A Brief study was performed which shows that max pooling performance in our image datasets is better than that of mean pooling layers. This article also supports that architecture with max-pooling layers outperforms those with mean pooling layers. Thus, all pooling layers which we used in our model is the max-pooling layers.

## 2.3 Activation Layers

The typical way of giving a non-linearity to our ANN is using sigmoidal functions and tanh nonlinearities, but these saturating non-linearity slows down the computations. Recently, the ReLU was introduced (Nair and Hinton, 2010) as a nonlinear activation function [2]. The advantage of using ReLU over other nonlinear activation functions is that it is not bounded to output values except to their negative values whereas, other nonlinear activations either gets saturated or are bounded to the output values or both. Intuitively the gradients of the ReLU takes only two values, i.e. either zeros or ones. These features of the ReLU makes the computations much faster than those using sigmoidal nonlinearities, and it also increases the efficiency and performance of the model.

## 2.4 Auxillary Layers

Overfitting has been a major issue in the field of machine learning. It is described as a phenomenon in which our model classifies our training dataset effectively but fails to classify over validation and testing results. To solve this issue, dropouts are often used. Training a network with a large amount of neurons often results in overfitting due to complex adaptations. Dropout is used to randomly disconnect connections between neurons of connected layers provided with a particular dropout rate value. Accordingly, the network can generalize training examples more efficient and effectively by reducing these adaptations. A trick was used for applying the dropouts, which takes average values of training dataset, which often reduces the network training time. Though the distribution of input layers shifts by passing through layers, which is also known as internal covariate shift, and has been pointed out as being a major point for slowing the training pace of the model. So to increase the training speed, whitening was performed. This technique causes the learning rate to increase which leads to much faster network convergence.

## 2.5 Softmax Layers

For the classification of data, a layer is required which may help in predicting data into different classes. This layer is a fully-connected layer and is situated at the last of our CNN architecture. The most prominent method of predicting classes is to use the softmax function given by equation (2), which is expresses as probabilistic function expression $p(y^{(i)} = n \mid x^{(i)} ;W)$ for the ith training example out of m number of training examples, the jth class out of n number of classes, and weights W, where $W_n Tx(i)$ are inputs of the softmax layer [2]. The sum of the right-hand side for the ith input always returns as 2, as the function always normalizes the distribution. In other words, Equation (2) returns the probabilities of each input's classes.

$$P(y^{(i)} = n | x^{(i)}; W) = \begin{pmatrix} P(y^{(i)} = 1 | x^{(i)}; W) \\ P(y^{(i)} = 2 | x^{(i)}; W) \\ \vdots \\ P(y^{(i)} = 3 | x^{(i)}; W) \end{pmatrix} \quad (2)$$

$$= \frac{1}{\sum_{j=1}^{n} e^{W_j^T x^{(i)}}} \begin{pmatrix} e^{W_1^T x^{(i)}} \\ e^{W_2^T x^{(i)}} \\ \vdots \\ e^{W_n^T x^{(i)}} \end{pmatrix} \quad (3)$$

### 2.6 CNN architecture

The CNN architecture which we used was VGG16 on very high-resolution images. The architecture consists of 16 layers with 4 convolution blocks. Last 4 layers of our mode were fully-connected Dense Layers. The last layer which we used was a softmax activation layer thus classifying our image into 4 classes, such that softmax function gives the probability of all 4 classes. The activation function which we used along with Conv-2D layers was ReLU, because it does not saturate, makes computations faster and is not bounded by outputs for positive values. The ReLU activation function performance on our datasets was much better than sigmoid and tanh nonlinearity activation functions. The weights which we used in our model were already pre-trained on ImageNet dataset, and we did not include the top layer of the VGG16. The input shape of all the images was 224 X 224 X 3. The input images consist of three channels, i.e., red, green and blue. For preventing overfitting of data, we used the dropout layers with a dropout rate of 0.5. Dropout layers were used after each Dense layers so as our model may give better performance on validation and test sets. The Optimizer which we used was the Stochastic Gradient Descent Algorithm with a learning rate of 1e-5 with a momentum value of 0.9. Stochastic Gradient Descent Algorithm is a bit slow as compared to Adam's Optimizer Algorithm but gave more accurate results than other Optimizer Algorithms. The Loss function which we used was Categorical Cross-entropy, which gave the probabilities of each corresponding class. The total images on which we trained and validated our network were 1000 and 200. The total number of epochs on which we ran our transfer learning VGG16 model was 60. We divided the images into the batch-size of 20, thus performing 50 iterations of forwarding propagation and backward propagation in each epoch. Our model was trained on 60 epochs to decrease the variations in the Testing and Validation Accuracies and Losses respectively.

**Table 1** Architecture and Details for VGG16

| Conv net architecture | | | Output shape |
|---|---|---|---|
| Block | Layer(type) | Filter size(#) | VGG size |
| Input | Input image | - | (N, 3, 224, 224) |
| | Convolutional | 3 × 3 (64) | (N, 64, 224, 224) |
| | Convolutional | 3 × 3 (64) | (N, 64, 224, 224) |
| Conv block 1 | Max pooling | - | ($N$, 64, 112, 112) |
| Conv block 2 | Convolutional | 3 × 3 (128) | ($N$, 128, 112, 112) |
| | Convolutional | 3 × 3 (128) | ($N$, 128, 112, 112) |
| | Max pooling | - | ($N$, 128, 56, 56) |
| Conv block 3 | Convolutional | 3 × 3 (256) | ($N$, 256, 56, 56) |
| | Convolutional | 3 × 3 (256) | ($N$, 256, 56, 56) |
| | Convolutional | 3 × 3 (256) | ($N$, 256, 56, 56) |
| | Max pooling | - | ($N$, 256, 28, 28) |
| Conv block 4 | Convolutional | 3 × 3 (512) | ($N$, 512, 28, 28) |
| | Convolutional | 3 × 3 (512) | ($N$, 512, 28, 28) |
| | Convolutional | 3 × 3 (512) | ($N$, 512, 28, 28) |
| | Max pooling | - | ($N$, 512, 14, 14) |
| Conv block 5 | Convolutional | 3 × 3 (512) | ($N$, 512, 14, 14) |
| | Convolutional | 3 × 3 (512) | ($N$, 512, 14, 14) |
| | Convolutional | 3 × 3 (512) | ($N$, 512, 14, 14) |
| | Max pooling | - | ($N$, 512, 7, 7) |
| Fully connected layer | Flatten | - | ($N$, 25,088) |
| | Dense | - | ($N$, 512) |
| | Dropout (0.5) | - | ($N$, 512) |
| | Dense | - | ($N$, 256) |
| | Dropout (0.5) | - | ($N$, 256) |
| | Dense | - | ($N$, 4) |

### 3.0 Building Database

For this research, earthquake reconnaissance building images were collected. Total of 1200 images was collected from Haiti in 2010, Nepal earthquake in 2015, Taiwan earthquake in 2016, Ecuador earthquake in 2016 [16-20]. All of these images are available to the public in Datacenterhub and Design-safe [21-24].

This data was divided into four damage categories. No damage, Minor damage, Major damage, and Collapse. Two hundred fifty images of each category were selected for training and 50 images each for validation.

Table 2. Characteristics for each damage category

| Damage Categories | Characteristics |
|---|---|
| No Damage | No structural and non-structural damage |
| Minor Damage | Non-structural Damage such as cracks in non-load bearing walls |
| Major Damage | Both structural and non-structural damage but no collapse |
| Collapse | Structure Collapsed |

### 4.0 Results

The image Dataset that we had prepared for training and validation was implemented on VGG16 model using Transfer Learning. There was a total of 1000 training images and 200 validation images. Our model consisted of 16 layers of VGG16 model, 4 Dense layers and a softmax layer, determining the probability of each of these 4 classes of damage. We have trained our network with 60 epochs with each epoch consisting of 50 iterations. The model was trained using three different learning rates, i.e., 1e-4, 1e-5 and 3e-5. The training and Validation Accuracies and Losses are shown in the figure described below. Our model gave the best performance with a learning rate of 1e-5.

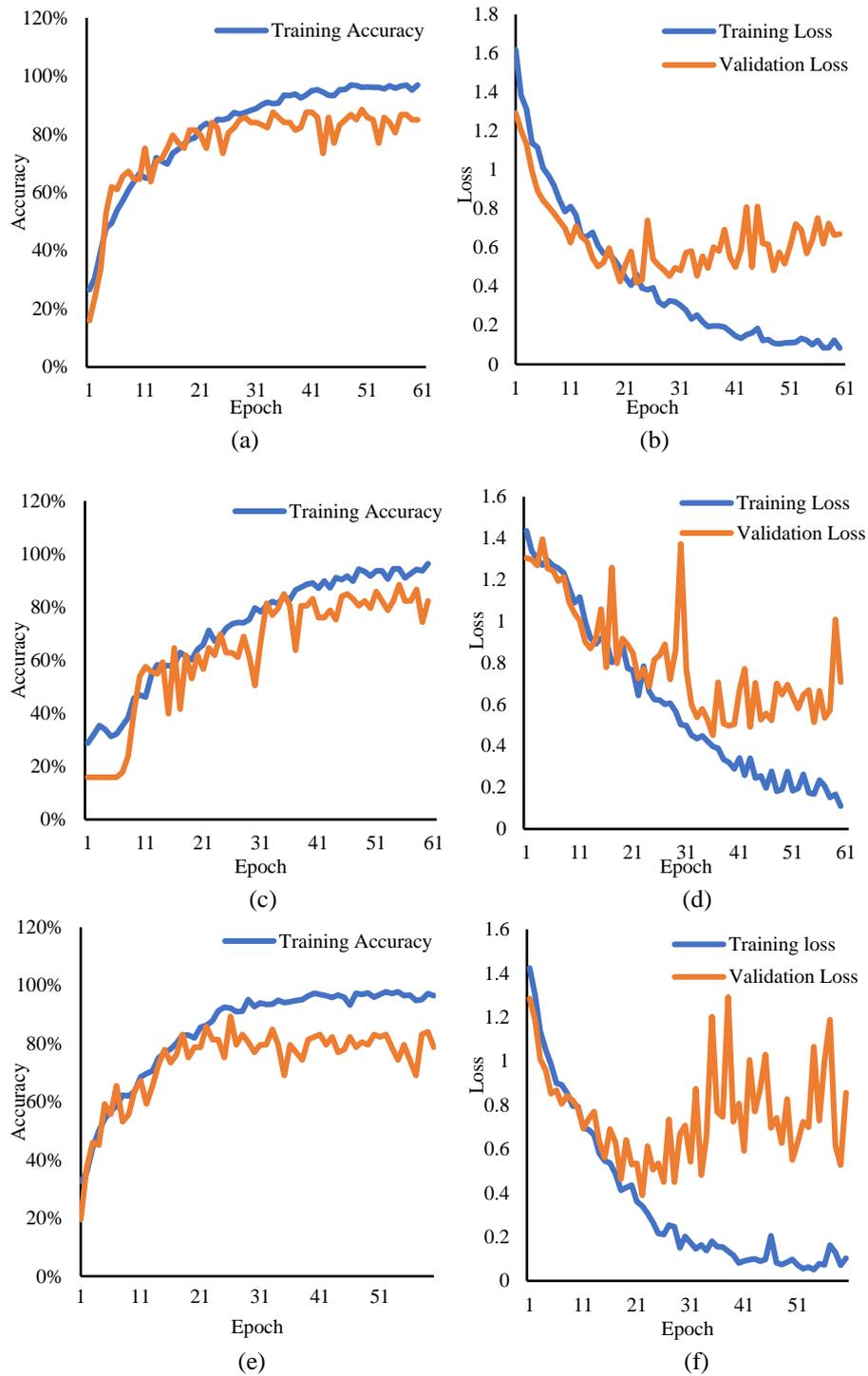

Figure 2: Accuracy history and loss of training and validation database at different learning rates (a) and (b) learning rate of 0.00001, (c) and (d) learning rate of 0.0001, (e) and (f) learning rate of 0.00003

The table shown below gives the training loss, training accuracy, validation loss and validation accuracy for each of the corresponding learning rates on which our model was trained.

**Table 2.** Loss and Accuracy for training and validation at different learning rates

| Learning Rates | Training Loss | Training Accuracy | Validation Loss | Validation Accuracy |
|---|---|---|---|---|
| 1e-5 | 0.0511 | 97.85% | 0.3893 | 89.38% |
| 3e-5 | 0.0838 | 96.91% | 0.4193 | 88.49% |
| 1e-4 | 0.1103 | 96.37% | 0.4512 | 88.49% |

**5.0    Conclusion**

The best results we got was using VGG16 transfer learning model. Our model gave a great performance on classifying our dataset to four damage levels as compared to AlexNet. The optimum loss and best accuracy we got on 60 epochs and 50 iterations. On training our model by further epochs, our training accuracy increased, but our validation accuracy started to decrease, thus showing that our model is under the stage of overfitting. Further, Dropout Layers were also used to prevent overfitting of our model. The difference in performance we got by using the Stochastic Gradient Descent Algorithm over Adam's Algorithm was about approximately 2% more; that is why we preferred the Stochastic Gradient Descent Algorithm. Our model also gave results of above 92% accuracy in classification of damages, when applied on test images.